\documentclass[1p,final]{elsarticle}
\makeatletter
\def\@author#1{\g@addto@macro\elsauthors{\normalsize%
    \def\baselinestretch{1}%
    \upshape\authorsep#1\unskip\textsuperscript{%
      \ifx\@fnmark\@empty\else\unskip\sep\@fnmark\let\sep=,\fi
      \ifx\@corref\@empty\else\unskip\sep\@corref\let\sep=,\fi
      }%
    \def\authorsep{\unskip,\space}%
    \global\let\@fnmark\@empty
    \global\let\@corref\@empty  %% Added
    \global\let\sep\@empty}%
    \@eadauthor={#1}
}
\makeatother

\usepackage{xytree}
\usepackage{comment}
\usepackage{amsmath,amssymb}
\usepackage{relsize}
\usepackage{subcaption}
\usepackage{hyperref}
\usepackage{setspace}
\journal{Journal of Information Processing and Management}
\DeclareMathOperator*{\argmax}{arg\,max}
\biboptions{authoryear}

\begin{document}

\begin{frontmatter}

\title{An enhanced Tree-LSTM architecture for sentence semantic modeling using \emph{typed dependencies}}

\author{Jeena Kleenankandy}\corref{mycorrespondingauthor}
\cortext[mycorrespondingauthor]{This is to indicate the corresponding author.}
\ead{jeenakk@gmail.com}
\author{ K A Abdul Nazeer}
\address{Department of Computer Science and Engineering}
\address{ National  Institute of Technology Calicut, Kerala-673601, India}

\begin{abstract}
\begin{spacing}{1.25}
\emph{Background}: Tree-based Long Short Term Memory (LSTM) network has become state-of-the-art for modeling the meaning of language texts as they can effectively exploit the grammatical syntax and thereby non-linear dependencies among words of the sentence. However, most of these models cannot recognize the difference in meaning caused by a change in semantic roles of words or phrases because they do not acknowledge the type of grammatical relations, also known as typed dependencies, in sentence structure.\\
\emph{Methods}: This paper proposes an enhanced  LSTM architecture, called \emph{relation gated LSTM}, which can model the relationship between two inputs of a sequence using a control input. We also introduce a Tree-LSTM model called \emph{Typed Dependency Tree-LSTM} that uses the sentence dependency parse structure as well as the dependency type to embed sentence meaning into a dense vector. \\
\emph{Results}: The proposed model outperformed its type-unaware counterpart in two typical Natural Language Processing (NLP) tasks -- Semantic Relatedness Scoring and Sentiment Analysis. The results were comparable or competitive with other state-of-the-art models. Qualitative analysis showed that changes in the voice of sentences had little effect on the model's predicted scores, while changes in nominal (noun) words had a more significant impact. The model recognized subtle semantic relationships in sentence pairs. The magnitudes of learned typed dependency embeddings were also in agreement with human intuitions.\\
\emph{Conclusion}:  The research findings imply the significance of grammatical relations in sentence modeling. The proposed models would serve as a base for future researches in this direction.
\end{spacing}
\end{abstract}

\begin{keyword}
Sentence Representation Learning \sep Universal Dependencies \sep Semantic Relatedness Scoring \sep Sentiment Analysis
%\MSC[2010] 00-01\sep  99-00
\end{keyword}

\end{frontmatter}

%\linenumbers

\section{Introduction}
\begin{figure}[h!]
  \centering
  \begin{subfigure}[b]{\linewidth}
    \includegraphics[width=0.8\linewidth]{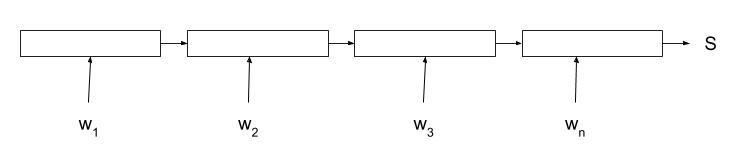}
    \caption{Sequential model for semantic composition}
  \end{subfigure}
  \begin{subfigure}[b]{\linewidth}
    \includegraphics[width=0.6\linewidth]{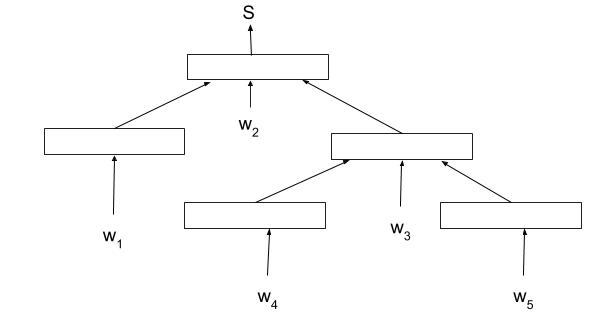}
    \caption{Tree-structured composition model based on dependency tree}
  \end{subfigure}
  \caption{Compositional model for sentence modeling. The $w_i$ are word embedding and $s$ is the semantic representation for the sentence}
  \label{fig:semcomp}
\end{figure}

Sentence modeling is a crucial step in Natural Language Processing (NLP) tasks, including but not limited to Sentence Classification \citep{liu2019bidirectional,xia2018novel}, Paraphrase Identification \citep{agarwal2018deep,jang2019recurrent}, Question Answering \citep{liu2019visual,zhu2020knowledge}, Sentiment Analysis \citep{kim-2014-convolutional,tai2015improved}, and Semantic Similarity Scoring \citep{shen2020learning,tien2019sentence}. Word meanings are represented using neural embedding \citep{bengio2003neural,mikolov2013efficient,pennington2014glove,turian2010word}, and sentence semantics are derived from these word vectors using compositional models. Many of the earlier models adopted for composition were either Bag-of-Words (BOW) \citep{mitchell2008vector} or sequential \citep{mueller2016siamese,tai2015improved}, as shown in Figure \ref{fig:semcomp}. The BOW model treats sentences as a mere collection of words, and compositional functions are simple vector operations \citep{mitchell2008vector}. Sequential models, on the other hand, consider text as a sequence of words. Most of the current deep learning sequential models for sentence semantic matching tasks use Long Short Term Memory (LSTM) \citep{wang2017bilateral} network or Convolutional Neural Network (CNN) \citep{kim-2014-convolutional,kalchbrenner-etal-2014-convolutional} or both \citep{agarwal2018deep,shakeel2020multi}. However, sequential models fail to capture non-linear dependencies between words that are common in natural languages.

Tree-structured models overcome this drawback by using model architectures that either reflects the parse tree of the sentences \citep{socher2010learning,socher2011parsing,socher2012semantic,socher2013recursive,tai2015improved} or latent trees learned for specific tasks \citep{choi2018learning,yogatama2016learning}. Tree-RNNs that use grammatical structures, have neural network units arranged along the nodes of a binarized constituency parse tree (CT) or dependency parse tree (DT). The key difference between these two architectures - CT-RNN and DT-RNN,  lies in their compositional parameters \citep{tai2015improved}. In CT-RNN, the word embeddings are fed only at leaf nodes. At every non-leaf node, the left and the right child nodes combine to form the parent node representation. In DT-RNN,  every node takes a word embedding and an unordered list of child nodes as compositional parameters.  \cite{tai2015improved} showed that DT-LSTM performs better in semantic relatedness scoring while CT-LSTM is more promising for sentiment classification. 

Though these Tree-based models effectively exploit the syntax and thus the relationships among words in a sentence, they ignore a valuable piece of information - the type of word relationship. A word-pair can have different types of grammatical relations in different sentences. These relationships are represented in the sentence's dependency tree. A dependency tree is a graphical notion of a sentence with each node corresponding to a word in the sentence. Each edge of a dependency tree, labeled by the dependency type, marks a dependency relation between the two words. Hereafter, we refer to these relations as Typed Dependencies\footnote{As in Stanford typed dependencies, which later evolved to Universal Dependencies. Both are used interchangeably in this paper.} \citep{de2008stanford,schuster2016enhanced}. These typed dependencies also contribute to the semantics of the text. 

\begin{figure}[h!]
  \centering
    \includegraphics[width=\linewidth]{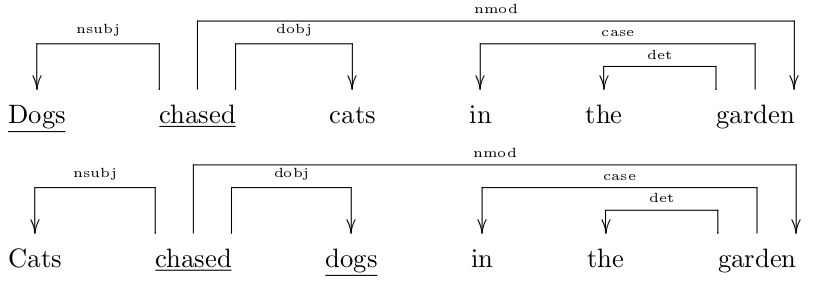}
  \caption{Two identical dependency trees that differ only in their edge labels. }
  \label{fig:trees}
\end{figure}

We illustrate this with an example. Consider the two sentences :
\begin{enumerate}[(1)]
 \item Dogs chased cats in the garden
 \item Cats chased dogs in the garden.
%\item Dogs were chased by the cat in the garden.
\end{enumerate}
In the sentences (1) and (2), the word-pair (Dogs, chased) shares a parent-child relationship in their dependency trees, as shown in Figure \ref{fig:trees}. 

 The typed dependency of (Dogs, chased) in the first sentence is ``dobj" for direct object whereas, in the second sentence, it is ``nsubj" for nominal subject. Nominal subject dependency indicates ``Dogs" is the subject for the action ``chased" while in the first case, direct object relation indicates ``Dogs" is the object of ``chased". A model that fails to acknowledge this difference treats both the relation as same.
 
The type of grammatical relation between words plays a crucial role in sentiment analysis, as well. The goal of sentiment classification is to detect the sentiment polarity of a given text and classify it usually as positive, negative, or neutral. Consider the phrases \emph{``white blood cells destroying an infection"} and \emph{``an infection destroying white blood cells"} \citep{socher2011semi}. Though they have the same words and tree structure, the former is positive, whereas the latter is very negative. A model that uses bag-of-words or parse structure without the dependency type can differentiate neither the semantics nor the sentiments of these phrases. Hence the need for a type-aware deep learning model.

Though some recent researches \citep{kim2019dynamic,liu2017dynamic,qian2015learning,wang2017tag}  try to address the inability of classic deep learning models to handle syntactic categories differently by using POS and constituency tags, the role of typed dependencies hasn't been much explored yet. The first notable work using typed dependencies was by \cite{socher2014tacl} in the Semantic Dependency Tree-RNN model (SDT-RNN). SDT-RNN is a recursive neural network modeled based on the dependency tree. The model trains a separate feed-forward neural network for each dependency type, making it highly data-intensive and complex to train. 

In Dependency-based Long Short-Term Memory (D-LSTM) network, \cite{10.1371/journal.pone.0193919} add an $\alpha$--weighted supporting component derived from the subject, object, and predicate of the sentence to the basic sentence representation generated by standard LSTM. The Part-of-Speech based LSTM (pos-LSTM) model by \cite{zhu2019part} derives the supporting component from the hidden representation of constituent words and its tag specific weights. The pos-LSTM model gave the best results when it used only the hidden representation of nouns in the sentence to compute the supporting component, which indicates noun words played a more significant role compared to others. The D-LSTM and pos-LSTM models are not syntax-aware as they are sequential, and the grammar is used only to identify semantic roles or POS tags. Nevertheless, these models showed improvement over their base model Manhattan LSTM (MaLSTM) \citep{mueller2016siamese}.

\cite{shen2020learning} proposed a Tag-Guided Hyper Tree-LSTM (TG-HTreeLSTM) model , which consists of a main Tree-LSTM and a hyper Tree-LSTM network. The hyper Tree-LSTM network uses a hypernetwork to dynamically predict parameters of main Tree-LSTM using the POS tag information at each node of the constituency parse tree. 
Structure-Aware Tag Augmented Tree-LSTM (SATA TreeLSTM) \citep{kim2019dynamic} uses an additional tag-level Tree-LSTM to provide supplementary syntactic information to its word-level Tree-LSTM. These models have shown remarkable improvement over the tag-unaware counterpart in sentiment analysis of sentences.
These facts motivated us to investigate further the role of grammar, precisely that of grammatical relationships in semantic composition.
%Tree-LSTMs\citep{tai2015improved,zhu2015long} are the most popular among tree-based models, with LSTM units arranged along a constituency or dependency parse tree. This work addresses the question, how can tree-based LSTM models be enhanced using typed dependencies to learn a better semantic representation of sentences.

\subsection{Research Objectives and Contributions}
This research has two primary objectives:
\begin{enumerate}[(1)]
 \item To propose a generic LSTM architecture that can capture the type of relationship between elements of the input sequence and, 
 \item To develop a deep neural network model that can learn a better semantic representation of sentences using its dependency parse structure as well as the typed dependencies between words. 
\end{enumerate}
This work also aims to perform a qualitative analysis of the results to study the role of typed dependencies in measuring semantic relatedness.

To capture relationship type, we introduce an additional neural network called “Relation gate” to the LSTM architecture that can regulate the information propagating from one LSTM unit to another based on an additional control parameter \emph{relation}. We use this “Relation gated LSTMs” to propose the Typed Dependency Tree-LSTMs for computing sentence representation. Our model is based on \cite{tai2015improved}'s Dependency Tree-LSTM and outperforms it in two sub-tasks - Semantic relatedness scoring and Sentiment Analysis.\\ %To the best of our knowledge, this is the first work in Tree-LSTMs that uses typed dependencies for semantic composition.\\
The contributions of this paper are:
\begin{enumerate}[(1)]
 \item Relation gated LSTM (R-LSTM) architecture that uses an additional control input to regulate the LSTM hidden state
 \item Typed Dependency Tree-LSTM model using R-LSTMs for learning sentence semantic representation over the dependency parse tree.
 \item A qualitative analysis of the role of typed dependencies in language understanding
\end{enumerate}
The rest of the paper is organized as follows. Section \ref{sec:background} gives a brief overview of LSTMs and Dependency Tree-LSTM architecture. The architecture of the proposed Relation gated LSTM (R-LSTM) and a formal description of the Typed DT-LSTM model are given in sections \ref{sec:rlstm} and \ref{sec:tydtree} respectively. The experiments are detailed in section \ref{sec:exp}. We report the results of these experiments and their qualitative analysis in section \ref{sec:dis}. Section \ref{sec:con} concludes the work with a remark on future prospects of the proposed model.

\section{Background}{\label{sec:background}}

\subsection{Long Short Term Memory (LSTM)}{\label{sec:lstm}}

\begin{figure}
  \centering
  \includegraphics[width=\linewidth]{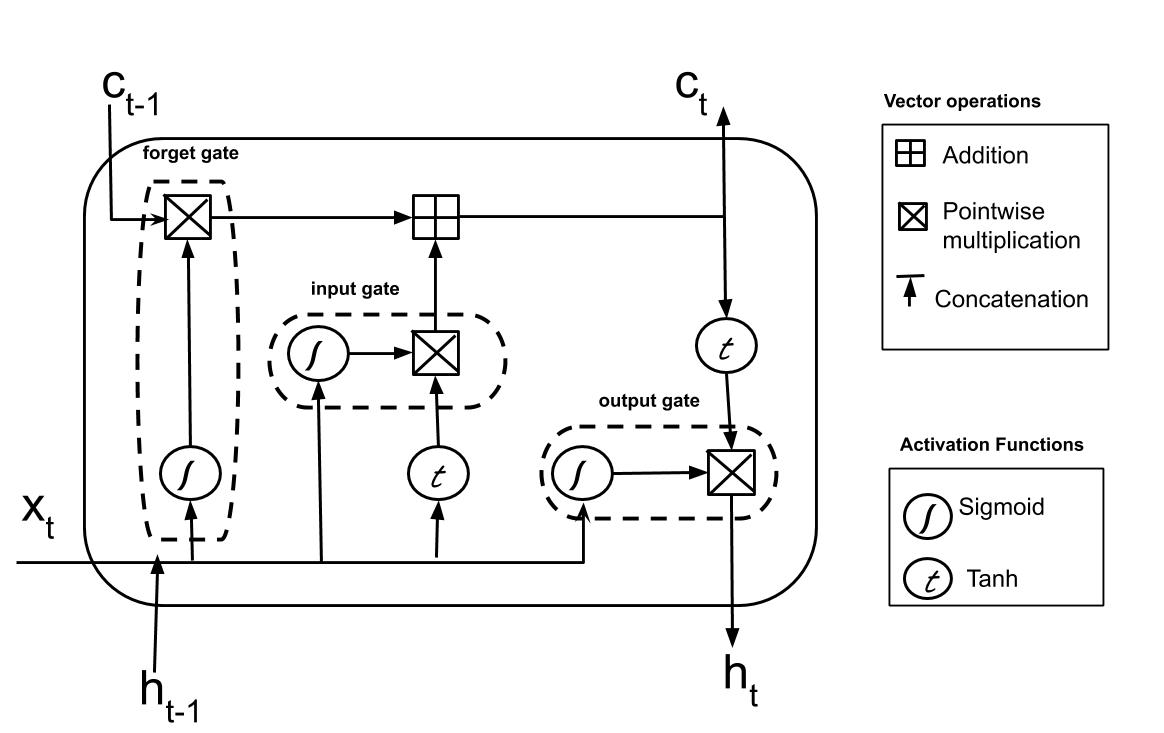}
  \caption{Architecture of Long Short Term Memory Unit \citep{hochreiter1997long}}
  \label{fig:lstm}
\end{figure}

Recurrent neural networks (RNN) can process input sequences of arbitrary length by repeatedly applying a transition function on each element of the sequence. The input to an RNN unit at time step $t$ is the current input $x_t$ and the hidden state $h_{t-1}$ from its previous time step. The final output of RNN could be either the output generated from the hidden state at the last time step or a sequence of outputs, one each from every hidden state, depending on the task addressed. 
LSTMs \citep{hochreiter1997long} are a form of RNN that can retain its state for longer sequences using its cell memory. Gates of an LSTM allow it to update the cell memory selectively. An LSTM unit consists of three gates -- forget gate, input gate and output gate, and two memory states -- cell state and hidden state. All these are vectors in $\mathbb{R}^d$, where $d$ is a hyperparameter -- the memory state dimension. Figure \ref{fig:lstm} shows the architecture of standard LSTM unit. Equation \ref{eqn:lstm} denotes the set of transitions of an LSTM unit at time step t.
\begin{equation}
\label{eqn:lstm}
\begin{aligned}
 i_t &= \sigma ( W^{(i)}x_t + U^{(i)}h_{t-1} + b^{(i)}),\\
 f_t &= \sigma ( W^{(f)}x_t + U^{(f)}h_{t-1} + b^{(f)}),\\
 o_t &= \sigma ( W^{(o)}x_t + U^{(o)}h_{t-1} + b^{(o)}),\\
 u_t &= tanh ( W^{(u)}x_t + U^{(u)}h_{t-1} + b^{(u)}),\\
 c_t &= i_t \odot u_t + f_t \odot c_{t-1},\\
 h_t &= o_t \odot tanh(c_t)
\end{aligned}
\end{equation}
where $\sigma$ denotes the logistic sigmoid function and $\odot$ is pointwise multiplication. The gating vector $f_t$, $i_t$, and $o_t$ are outputs of feedforward neural networks that take $x_t$ and previous hidden state $h_{t-1}$ as inputs. The forget gate $f_t$ restrains the extent of information propagated from the previous hidden state. The input gate $i_t$ controls what information from the current input supplements the cell state. The output gate $o_t$ is responsible for selecting how much of the cell state flows to the next time step as its hidden state. By using these three gates, LSTM units are capable of selecting only vital information from the inputs, retaining and propagating them along the LSTM chain. Standard LSTMs are strictly sequential and hence unable to acknowledge any non-linear dependencies in the input sequence.

\subsection{Dependency Tree-LSTMs}\label{sec:dtlstm}

Tree-structured models are a natural choice for text processing as the sentences have an inherent hierarchical structure conveyed by its parse tree. A pioneering work in this category is the Tree-LSTM, proposed by \cite{tai2015improved}, in which each LSTM unit can take input from multiple LSTM units forming a tree topology. \cite{tai2015improved} also proposed two variants of Tree-LSTM, namely - Child-Sum Tree-LSTMs and N-ary Tree-LSTMs for sentence representation. The former is suited for trees having nodes with an arbitrary number of unordered children, while the latter is for trees having nodes with at most N children in a fixed order. A child-sum tree-LSTM over dependency parse tree is known as Dependency Tree-LSTM (DT-LSTM). 
\begin{figure}
  \centering
  \includegraphics[width=0.5\linewidth]{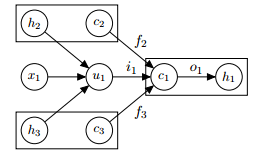}
  \caption{Composition of a Dependency Tree-LSTM node with two children \citep{tai2015improved}}
  \label{fig:childsum}
\end{figure}
 Each node of a DT-LSTM takes as input a vector $x_t$ and an arbitrary number of hidden states, one from each of its child nodes. Equation \ref{eqn:dtlstm} denotes the set of transitions of a DT- LSTM unit at time step t.  
%\begin{equation}

\begin{align}
 h_{C(t)} &= \displaystyle\sum_{k\in C(t)} h_k,\nonumber\\
 i_t &= \sigma ( W^{(i)}x_t + U^{(i)}h_{C(t)} + b^{(i)}),\nonumber\\
 f_{tk} &= \sigma ( W^{(f)}x_t + U^{(f)}h_{k} + b^{(f)}), \; k \in C(t)\nonumber\\
 o_t &= \sigma ( W^{(o)}x_t + U^{(o)}h_{C(t)} + b^{(o)}),\label{eqn:dtlstm}\\
 u_t &= tanh ( W^{(u)}x_t + U^{(u)}h_{C(t)} + b^{(u)}),\nonumber\\
 c_t &= i_t \odot u_t + \displaystyle\sum_{k\in C(t)} f_{tk} \odot c_{k},\nonumber\\
 h_t &= o_t \odot tanh(c_t)\nonumber
\end{align}
%\end{equation}
where $C(t)$ is the set of all child nodes of the node $t$.

Unlike the standard LSTMs, the gating vectors and cell states of the DT-LSTM unit depend on the hidden state of not just one but multiple child units. The DT-LSTM unit has multiple forget gates, one for each of its child units, allowing it to emphasize information from selected child nodes based on the task addressed.  At each node $t$ of the DT-LSTM, the input vector $x_t$ is the word embedding corresponding to the headword at that node. The hidden state $h_t$ is an abstract representation of the sub-tree rooted at $t$. The cell state of the parent node depends on the sum of the cell states of its child nodes, hence the name child-sum LSTMs (Refer Figure \ref{fig:childsum}).

Though Tree-based LSTMs respect non-linear dependencies between words in the sentence, neither Tree-LSTMs nor any of its variants explicitly make use of the type of these dependencies. The intuition behind our proposed Typed Dependency Tree-LSTM is that an LSTM unit can learn to make informed decisions on what information it passes to the next LSTM unit if it knows the kind of dependency they share. To equip LSTM with this knowledge, we propose the relation gated LSTM (R-LSTM) with an additional control input $z_t$. R- LSTM has an additional gate, hereafter referred to as relation gate $r_t$, that can regulate its hidden state $h_t$ based on this control input $z_t$. The proposed architecture is discussed in section \ref{sec:rlstm}. 

\section{Relation gated LSTM Architecture}{\label{sec:rlstm}}
Relation gated LSTM (R-LSTM) is an enhancement of standard LSTM. 
\begin{figure}
  \includegraphics[width=\linewidth]{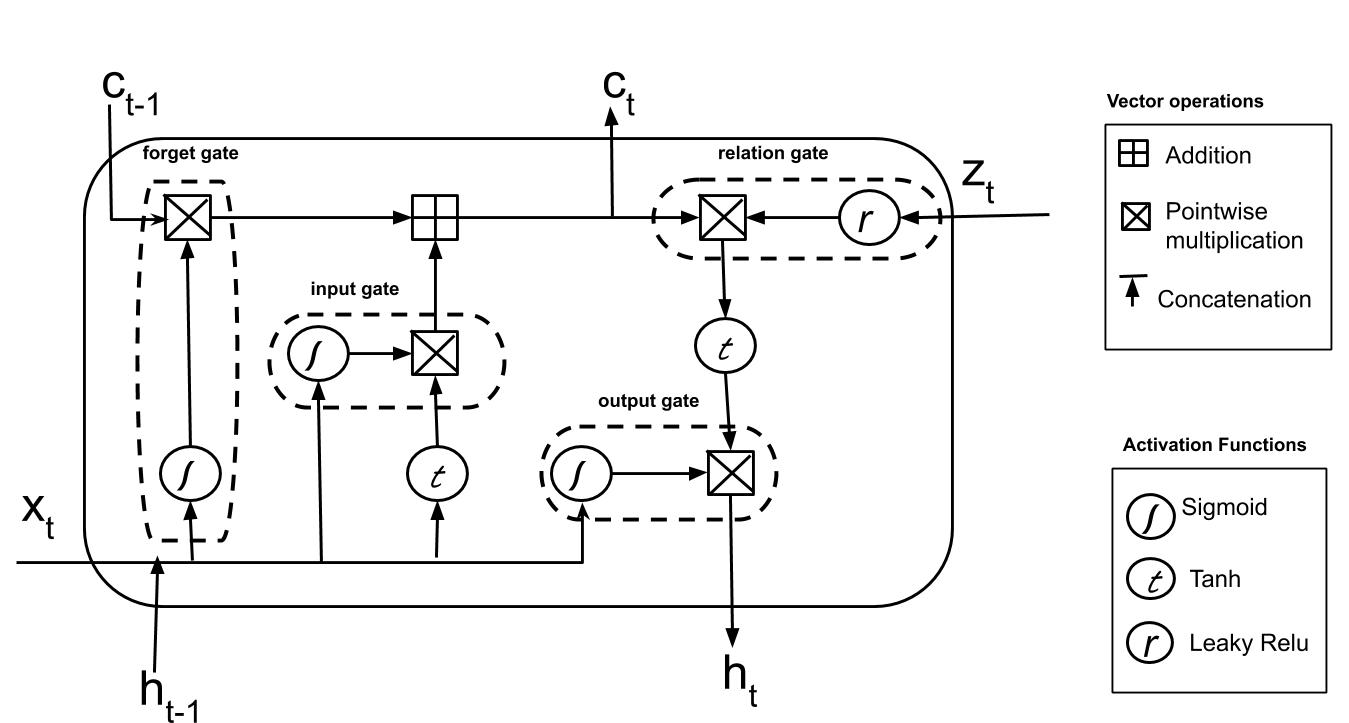}
  \caption{Architecture of Relation gated Long Short Term Memory Unit}
  \label{fig:rlstm}
\end{figure}
R-LSTM consists of 4 gates -- forget gate, input gate, output gate and a relation gate, and two memory states -- the hidden state and the cell state. The architecture of R-LSTM is shown in figure \ref{fig:rlstm}. The relation gate and hidden state of R-LSTM at time step $t$ with successor $t'$ is given by the equation \ref{eqn:rlstm}. For sequential model $t^{'}=t+1$ whereas for tree-based model $t^{'}$  depends on the tree topology used. 
\begin{equation}
\label{eqn:rlstm}
\begin{aligned}
 r_{tt'} &= g ( W^{(r)}z_{tt'} + b^{(r)}),\\
 h_{tt'} &= o_t \odot tanh(c_t \odot r_{tt'})
\end{aligned}
\end{equation}
where $g$ is a non-linear activation function.

The three gates $i_t$, $f_t$, $o_t$ and other vectors are computed using equation \ref{eqn:lstm} of standard LSTM (Refer Section \ref{sec:lstm}). The relation gating vector $r_{tt^{'}}$ controls how much information from the cell state $c_t$ should be transferred to the hidden state $h_t$ of unit $t$ and thereby propagate to the next unit $t^{'}$. The control input $z_{tt'}$ represents the relation between the inputs $x_t$ and $x_t'$. The relation gate of R-LSTM depends only on this control input $z_{tt'}$. Unlike standard LSTMs, the hidden state of R-LSTM depends not just on the cell state and the output gate but also on the relation gate.

\begin{figure}[h!]
  \centering
    \includegraphics[width=\linewidth]{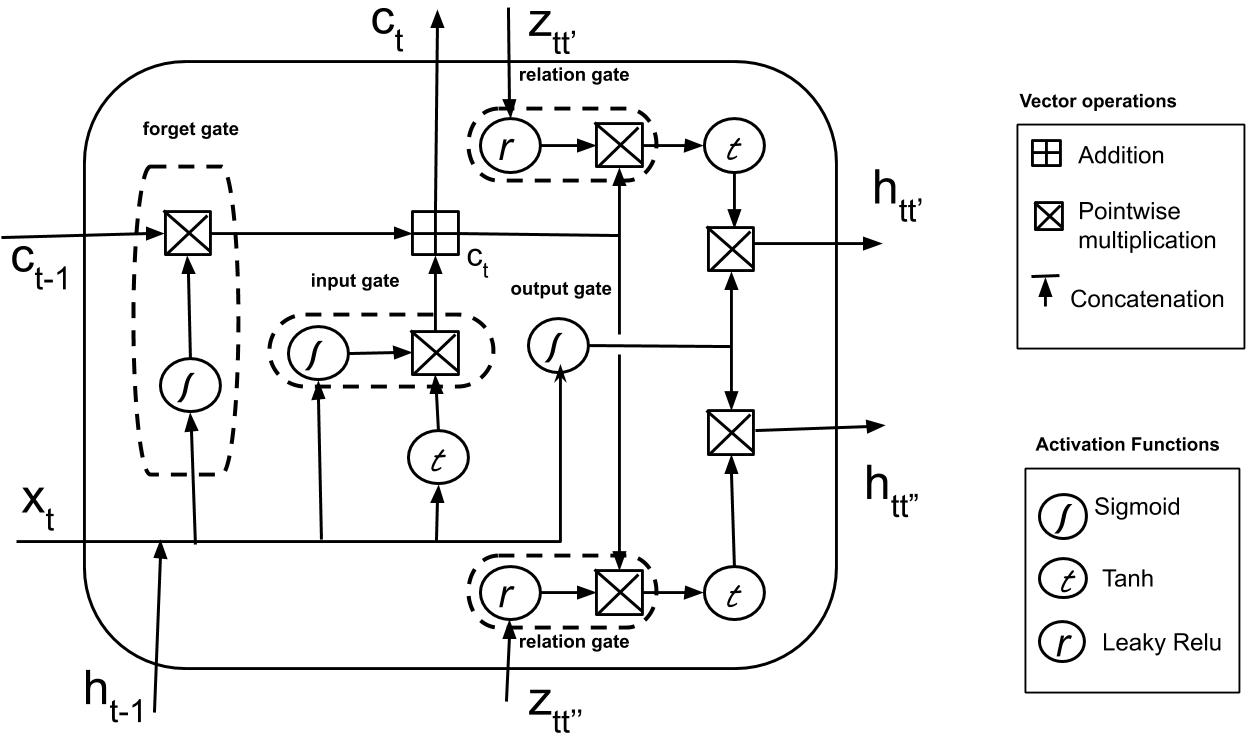}
    \caption{R-LSTM with two control inputs and two hidden states.  }
    \label{fig:multirlstm}
\end{figure}

 Relation gate is useful in the NLP scenario where words can share different types of relations, and these relation types affect their semantic composition. In our proposed Typed DT-LSTM, the control input $z_t$ represents the type of dependency the word at node $t$ has with its headword at node $t^{'}$. Note that, in a dependency parse tree, any node $t$ can have only one parent $t^{'}$. In the Dependency Acyclic Graphs (DAGs) representation, a word can have different types of dependencies with different words in the sentence, i.e., a node can have more than one parent. Accommodating such multiple relationships is trivial in R-LSTMs. In such cases, R-LSTM would have multiple control inputs and therefore, multiple hidden states, one for each of its dependencies. Having multiple hidden states allow an R-LSTM unit to transmit different information to each of its parent units based on the kind of relation it shares. Figure \ref{fig:multirlstm} shows an R-LSTM unit with two parent units. The relation gates are drawn separately for clarity, though R-LSTM unit has only one relation gate and thus only one relation weight matrix $W^{(r)}$. The relation gating vector $r_{tt'}$ can be computed by equation \ref{eqn:rlstm} for each $t' \in P(t)$, parents of node $t$.
\section{Typed Dependency Tree-LSTMs}\label{sec:tydtree}
As mentioned in section \ref{sec:dtlstm} , Dependency Tree-LSTM uses the same set of weights for every LSTM units and hence is not aware of the relation type between them. In this section, we propose Typed DT-LSTM model that can learn embeddings of the type of grammatical relation between the word pairs using R-LSTMs. We hypothesize that this knowledge helps our model to built better semantic representation of sentences. Like the DT-LSTM \citep{tai2015improved}, our Typed DT-LSTM architecture also follows the dependency tree topology. In addition, we make use of the typed dependencies obtained by the dependency parse, which is detailed below.\\
Let $D$ be a ordered universal set of typed dependencies in the language.
\begin{equation}
D = [d_1, d_2,... d_l] , l = |D|  
\end{equation}
Given a sentence $S = (w_1,w_2,...,w_n)$ with $n$ words. The dependency parse of $S$ is defined by a set of typed dependencies, 
\begin{equation}
TD(S) = \{d_j(w_{t^{'}},w_t), d_j \in D, w_t \in S, w_{t^{'}} \in S \cup \{ROOT \} \} 
\end{equation}
$TD(S)$ directly maps onto a dependency tree with nodes as words (except the root node) and edges labeled $d_j$ from node $w_t$ to $w_{t^{'}}$ for every $d_j(w_{t^{'}},w_t)$ in $TD(S)$ . The root of the tree is a special node $ROOT$. An edge labeled $root$ connects the root word of the sentence to the $ROOT$.
\begin{figure}[h!]
  \centering
    \includegraphics[width=\linewidth]{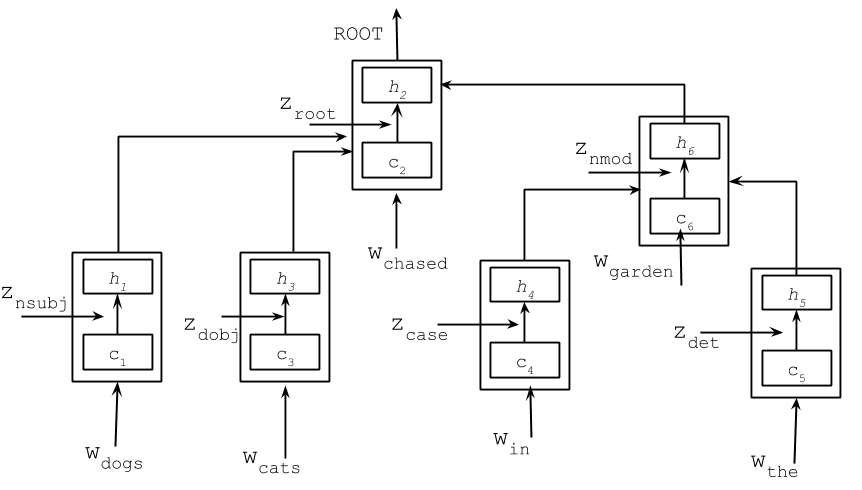}
    \caption{Semantic composition using Typed Dependency Tree-LSTM. z is the one-hot encoding of typed dependency and w is pre-trained word embedding. Each node is an R-LSTM unit. }
  \label{fig:tydtreelstm}
\end{figure}
 Each R-LSTM$^{(t)}$ unit in the model corresponds to a node $t$ of the tree. The R-LSTM$^{(t)}$ takes as input -- (1) $x_t$, the vector representation of the word $w_{t}$, (2) the relation vector $z_t = e^{(j)}$ ( a binary vector with $1$ at dimension $j$ and $0$ elsewhere) , if $d_j(w_{t^{'}},w_t) \in TD(S) $  and (3) the output from its child R-LSTM units. The output of R-LSTM$^{(t)}$ -- i.e. the cell state $c_t$\fnref{ztt} and hidden state $h_t$, propagates to its parent R-LSTM$^{t^{'}}$.
 \fntext[ztt]{In dependency tree,any node $t$ can have only one parent, so we can drop the subscript $t'$}
 We can now combine the equations \ref{eqn:dtlstm} and \ref{eqn:rlstm}, to formally define the Typed DT-LSTM as given in equation \ref{eqn:tydtree}
%\begin{equation}
\begin{align}
 h_{c(t)} &= \displaystyle\sum_{k\in C(t)} h_k,\nonumber\\
 i_t &= \sigma ( W^{(i)}x_t + U^{(i)}h_{c(t)} + b^{(i)}),\nonumber\\
 f_{tk} &= \sigma ( W^{(f)}x_t + U^{(f)}h_{k} + b^{(f)}),\nonumber\\
 o_t &= \sigma ( W^{(o)}x_t + U^{(o)}h_{c(t)} + b^{(o)}),\label{eqn:tydtree}\\
 u_t &= tanh ( W^{(u)}x_t + U^{(u)}h_{c(t)} + b^{(u)}),\nonumber\\
 c_t &= i_t \odot u_t + \displaystyle\sum_{k\in C(t)} f_{tk} \odot c_{k},\nonumber\\
 r_t &= g ( W^{(r)}z_t + b^{(r)}),\nonumber\\
 h_t &= o_t \odot tanh(c_t \odot r_t)\nonumber
\end{align}
%\end{equation}
The relation gating vector of the node $t$ given by $r_t \in \mathbb{R}^d$ is learned using $z_t \in \mathbb{R}^l$ the one-hot encoding of the typed dependency between $t$ and its parent node\footnote{subscript t' is dropped because a node can have only a single parent in a tree} in the dependency tree.

 Intuitively, the weight matrix $W^{(r)} \in \mathbb{R}^{l\times d}$ can be interpreted as a task-specific embeddings of typed dependencies, where each column $w_j$ corresponds to the dependency type $d_j$ in the set $D$.  This typed dependency embedddings are analyzed in detail in section \ref{sec:tdembd}.
 
\section{Experiments}\label{sec:exp}

We evaluate our model, Typed DT-LSTM using relation gated LSTM, on two tasks : (1)  Semantic relatedness scoring of sentence pairs and (2) Sentiment classification of movie reviews. 

For both the tasks, we follow the training and evaluation method proposed by \cite{tai2015improved}. The sentences in the training set are parsed using Stanford dependency parser \citep{chen2014fast,manning-EtAl:2014:P14-5}. We initialize word representations with pre-trained Glove \citep{pennington2014glove} word embeddings\footnote{Glove : \url{http://nlp.stanford.edu/data/glove.42B.300d.zip}}. All parameters are randomly initialized. The 47 universal typed dependencies are encoded as one-hot vectors. The proposed Typed DT-LSTM generates a sentence embedding for each sentence. We use a softmax classifier to generate the semantic score/sentiment label from this sentence semantic vector(s). Supervised training algorithm trains the model end-to-end to generate sentence representation as well as semantic score/sentiment labels. For semantic relatedness scoring, the word embeddings are kept fixed; they are not learnable parameters of the model.

\subsection{Semantic Relatedness Scoring}
Semantic relatedness scoring aims to predict a real-valued relatedness score for a given sentence pair based on the semantic similarity between its sentences. For each pair of a sentence ${(S_l, S_r)}$ in the training set, the Typed DT-LSTM model generates a sentence vector pair $(h_l,h_r)$ . We follow the Siamese architecture \citep{bromley1993signature} where the model weights for sentences in the pair are tied. The element-wise product $(h_{l} \odot h_{r})$ and absolute difference $(|h_l - h_r|)$ is then input to a neural network classifier. The classifier uses the softmax function to predict the semantic similarity of sentences as a probability distribution $\hat{p}$ over the $K$ classes. The equation \ref{eqn:srs} calculates the predicted semantic score,
\begin{eqnarray}
\begin{aligned}
 h_s &= \sigma \mathlarger{(} U(h_{l} \odot h_{r}) + V (|h_l - h_r|) + b_h \mathlarger{)},\\
 \hat{p}_\theta &= softmax(Wh_s + b_p),\\
 \hat{y} &=  r^T \hat{p}_\theta\; ,\; r = [1,2,3...,K]
\end{aligned}
\label{eqn:srs}
\end{eqnarray}
To train the model, we construct a target probability distribution $p$ from the actual similarity score $y$, such that $y = r^Tp$.
Each $i^{th}$ element $p_i$ in $p$ is assigned as, 
  \begin{equation}
  \label{eqn:srsp}
p_i =\begin{cases}
               1 - |i-y|,& i \in [\lfloor y \rfloor, \lceil y\rceil]\\
               0, & \text{otherwise}
            \end{cases}
  \end{equation}
The model trained by back-propagation minimize the regularized KL-divergence between $p^{(j)}$ and $\hat{p}^{(j)}_\theta$ for each sentence pair $(S_l, S_r)^{(j)}$ in the training set of size $m$.
\begin{equation}
J(\theta) = \frac{1}{m} \mathlarger{\displaystyle\sum_{j=1}^m} KL \left(p^{(j)}\|\hat{p}^{(j)}_\theta \right) + \frac{\lambda}{2}\|\theta\|^2_2,
\end{equation}
\subsection{Sentiment Classification}
We model sentiment classification as a binary classification task to classify a given sentence as positive or negative based on its sentiment polarity. The training data (Refer Section \ref{sec:dataset}) for the model is a set of dependency parse trees having some subset of its nodes annotated with sentiment labels. Each node label indicates the sentiment of the phrase spanned by that node. For instance, in the review \emph{``It's not a great monster movie"}, node spanning \emph{``a great monster movie"} is marked positive while that of \emph{``not a great monster movie"} is labeled negative. The review as a whole is labeled negative at the root node. The model aims to predict the root node's label recursively from its child nodes.

Typed DT-LSTM generates hidden representation for every node of the dependency parse trees in the training set. The hidden representation $h_t$ is an abstract representation of the sentiment of the phrase spanned by the node $t$.
At each node $t$, a softmax classifier maps the hidden representation $h_t$ to a probability distribution $p^{(t)}_{\theta}(y|\{x\}_t)$ over $K$ classes. The value of $p^{(t)}_{\theta}(y|\{x\}_t)$  is the probability of class label $y$ given $\{x\}_t$ -- the subtree rooted at the node $t$ .The predicted class label $\hat{y}^{(t)}$ of node $t$ is the class label $y$ with the maximum $p^{(t)}_{\theta}(y|\{x\}_t)$ value. 
%\begin{equation}
\begin{align}
  p^{(t)}_{\theta}(y|\{x\}_t) &= softmax(Wh_t + b),\\
  \hat{y}^{(t)} &= \argmax_{y}\;p^{(t)}_{\theta}(y|\{x\}_t).\nonumber
\end{align}
%\end{equation}
The supervised learning algorithm tries to minimize the negative log-likelihood of the true class labels $y^{(t)}$ at each labeled node:

\begin{equation}
J(\theta) = -\frac{1}{m} \mathlarger{\displaystyle\sum_{j=1}^m} \log \hat{p}^{(j)}_{\theta} \left(y^{(j)}\|\{x\}^{(j)} \right) + \frac{\lambda}{2}\|\theta\|^2_2,
\end{equation}
where $m$ is the total number of labelled nodes in the training set, and $\lambda$ is the L2 regularization parameter for the model.
\subsection{Datasets and Training details} \label{sec:dataset}
 For semantic relatedness scoring, we use Sentence Involving Compositional Knowledge (SICK)\citep{marelliraffaella}\footnote{SICK : \url{http://alt.qcri.org/semeval2014/task1/index.php?id=data-and-tools}} dataset consisting of 9927 sentence pairs each annotated with a real-valued score, in the range 1 -- 5 (1 being the least similar). The sentence pairs in the SICK dataset are sentences extracted from image and video description annotated by the average of ten human assigned scores. The original predefined train/dev/test split of 4500/500/4927 is used for the experiments.
 
 For Sentiment classification, we used the Stanford Sentiment Treebank (SST) \citep{socher2013recursive} dataset \footnote{SST dataset\url{http://nlp.stanford.edu/~socherr/stanfordSentimentTreebank.zip}}. The SST dataset consists of 11,855  sentences parsed using Stanford constituency parser to obtain 215,154 unique phrases. These phrases that correspond to nodes in the standardized binary constituency tree have manually assigned sentiment labels - positive, negative, or neutral. We generated dependency parse trees for these sentences. The dependency tree nodes whose node span matched the phrases in the dataset were labeled accordingly. Out of the original train/dev/test splits of 8544/1101/2210, after removing the neutral sentences 6920/872/1821 were used for binary classification. Dependency parse tree has a lesser number of nodes than its constituency counterpart, and not all nodes of the dependency tree have a matching phrase in the constituency parse tree. Thus we could label and use only a subset of the actual dataset for this experiment.
 
 The range of hyperparameters used is listed in table \ref{tab:hyperp}. The optimal hyperparameter values are chosen based on empirical results. Early stopping strategy was used for training with patience set to 10 epochs.
\begin{table}  
\begin{tabular}{@{}p{3cm}p{4cm}p{4cm}@{}}
\hline \hline
Parameter & SICK-R & SST \\\hline
Learning rate & 0.01/0.05/0.1/0.2/\textbf{0.25}/0.3\  & 0.01/\textbf{0.05}/0.1/0.2/0.25/0.3\\
Batch size  & \textbf{25}/50/100 & \textbf{25}/50/100\\
%Hidden dimension & \textbf{50}/100/150  & \textbf{50}/100/150 \\
Memory dimension & 120/\textbf{150}/100 & 165/168/\textbf{170}/175\\
Weight decay & 0.0001 & 0.0001\\
Optimizer & \textbf{adagrad}/adam/nadam & \textbf{adagrad}/adam/nadam\\
\hline
\end{tabular} 
\caption{Range of hyperparameters used for tuning the model. The best is shown in \textbf{bold}}
\label{tab:hyperp} 
\end{table}
\section{Results and Discussion}\label{sec:dis}
\subsection{Modeling semantic relatedness}
 For semantic relatedness scoring, we evaluated the model's performance using Pearson Correlation Coefficient and Mean-Squared Error between the actual relatedness score $y$ and the predicted score $\hat{y}$. We compare our results with deep learning models that use LSTMs or dependency tree or both for semantic composition. The models belong to four categories. The mean vector is the baseline, where the word vectors are averaged to get the sentence vector. In the first category we discuss sequential models that use LSTMs and GRUs (Gated Recurrent Units). In the second category are the tree-NN models that use the dependency tree structure but not the dependency type for composition. The third category consists of sequential models that use grammatical information to improve sentence representation. Our model falls in the fourth category, where the models use dependency structure along with the dependency type.
\begin{table}
\centering
\begin{tabular}{@{}p{4cm}p{2cm}p{2cm}@{}}
\hline \hline
Model & Pearson's $r$ & MSE \\\hline
Mean vectors & 0.7577  & 0.4557\\
\hline
Seq. LSTM  & 0.8528 & 0.2831\\
Seq. Bi--LSTM& 0.8567  & 0.2736 \\
Seq. GRU & 0.8595 & 0.2689\\
MaLSTM & 0.8177 & -- \\
\hline
DT-RNN & 0.7923 & 0.3848\\
DT-LSTM & 0.8676 & 0.2532\\
DT-GRU & 0.8672 & 0.2573\\
\hline
D-LSTM & 0.8270 & 0.3527\\
pos-LSTM-n & 0.8263 & -- \\
pos-LSTM-v & 0.8149 & -- \\
pos-LSTM-nv & 0.8221 & --\\
pos-LSTM-all & 0.8173 & -- \\
\hline 
SDT-RNN & 0.7900 & 0.3848\\
\textbf{Typed DT-LSTM} & \textbf{0.8731} & \textbf{0.2427}\\\hline
\end{tabular}
\caption{Comparison of Typed Dependency Tree-LSTM with other LSTM models for semantic relatedness scoring. The values are taken from previously published results.}
\label{tab:results}
\end{table}

Table \ref{tab:results} shows that LSTM models are more efficient than standard RNN models, due to LSTM's ability to retain information over longer sequences. The DT-LSTMs and DT-GRUs perform better than their sequential counterparts, validating that hierarchical structures are better in representing sentence semantics than sequences.

 The D-LSTM model and pos-LSTM models that use word's grammatical roles as additional information shows improvement over MaLSTM. For a fair comparison, we used MaLSTM and D-LSTM models without regression calibration, synonym augmentation, and pretraining \citep{10.1371/journal.pone.0193919}. The pos-LSTM-n that uses only noun words performs better than the pos-LSTM model that uses nouns and verbs, or all words. The D-LSTM that uses only the subject, predicate, and object fares better than all pos-LSTM models. These results indicate that words in certain grammatical roles are more useful in semantic representation.

In the fourth category, the only other model we are aware of that uses the dependency type is Semantic Dependency Tree-RNN model by \cite{socher2014tacl}. The SDT-RNN, which was initially proposed for image captioning, does not show improvement over DT-RNN in relatedness scoring tasks. The major concern of the model is its complexity and network size. In SDT-RNN, each dependency type is represented by a separate neural network with a weight matrix of size $2d \times d$ where $d$ is the dimension of the word embedding. With $300$ dimension GloVe embedding and 47 types of Universal Dependencies, the number of parameters for the network is $600 \times 300 \times 47 $. The SICK dataset is insufficient to train such a huge network efficiently. Typed DT-LSTM model represents each dependency type by a vector of $150$ (size of the memory state) dimension and thus has a lesser number of parameters to be trained. 
%To the best of our knowledge, Typed DT-LSTM is the first deep learning model that is effectively able to use the dependency type of word relations for sentence modeling.
\begin{figure*}[t!]
    \centering
    \begin{subfigure}[t]{0.45\textwidth}
        \centering
        \includegraphics[scale=0.35]{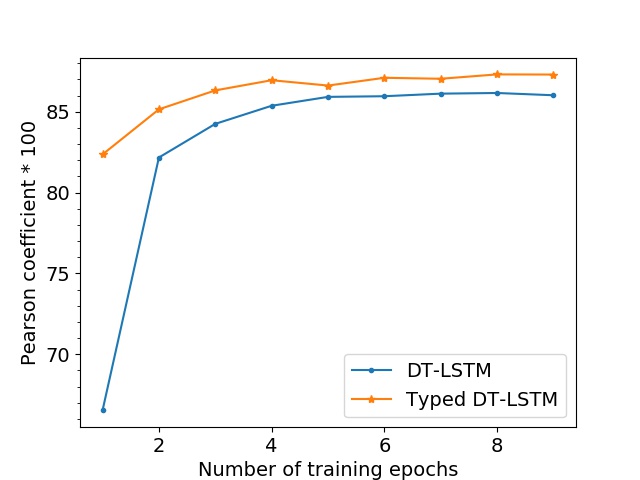}
        \caption{Pearson Correlation Coefficient}
    \end{subfigure}%
    \begin{subfigure}[t]{0.5\textwidth}
        \centering
        \includegraphics[scale=0.35]{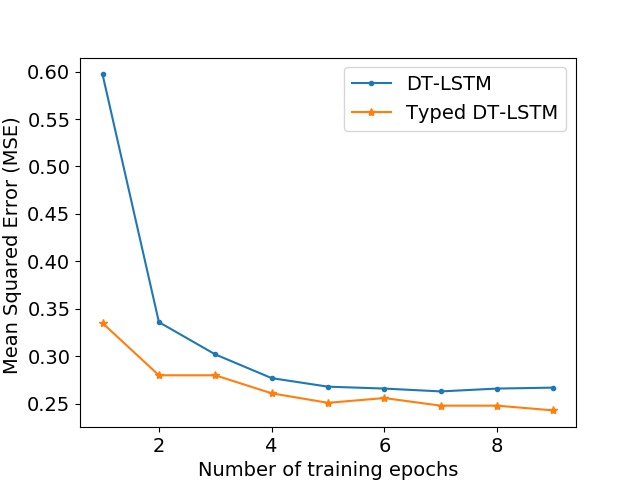}
        \caption{Mean Squared Error (MSE)}
    \end{subfigure}
    \caption{Comparison of learning curves of DT-LSTM and Typed DT-LSTM model for Semantic Similarity Task using (a) Pearson Correlation Coefficient and (b) Mean Squared Error (MSE) between the actual relatedness score $y$ and the predicted score $\hat{y}$. Learning curves are calculated from SICK Test dataset.}
    \label{fig:curves}
\end{figure*}
We also compared the learning curves of Typed DT-LSTM with that of the base model DT-LSTM, as shown in Figure \ref{fig:curves}. The experiments show that Typed DT-LSTM can learn faster than DT-LSTM, i.e., with a fewer number of training epochs.  As the only upper hand Typed DT-LSTM has over DT-LSTM is the knowledge of dependency types, these results validate our hypothesis: typed dependencies can assist Tree-LSTM to learn semantic representations.

\begin{table} 
\centering
\begin{tabular}{@{}p{8cm}p{1.5cm}p{1.5cm}@{}}
\hline \hline
Query and retrieved sentences & $S_{DT}$ & $S_{Typed-DT}$\\
\hline
\textbf{The turtle is following the fish} &  & \\
The turtle is following the fish & 4.48 & 4.81\\
The turtle is following the red fish & 4.48 & 4.66 \\
The fish is following the turtle & 4.48 & 4.57 \\
\hline
\textbf{A boy is running towards the ocean} &  & \\
A boy is running through the sand & 4.12 & 4.51\\
The young boy is running through the ocean waves & 4.11 & 4.15 \\
A boy is running towards the fishing line & 4.16 & 4.11\\
\hline
\textbf{A guy is riding a horse} &  & \\
A guy is riding a horse & 4.79 & 4.87\\
A horse is being ridden by a guy & 4.82 & 4.88 \\
The horse is being ridden by a man & 4.85 & 4.87\\
\hline
\end{tabular}
\caption{Three most similar sentences retrieved by Typed DT-LSTM from the SICK test set for each query sentence. $S_{DT}$ is the score assigned by Dependency Tree-LSTM and $S_{Typed-DT}$ is the score by the proposed Typed DT-LSTM.}
\label{tab:exrank}
\end{table}

Table \ref{tab:exrank} shows the three most similar sentences retrieved from the SICK test dataset for three sample queries. For the first query \emph{``The turtle is following the fish"}, the DT-LSTM assigns equal scores for both sentences, \emph{``The turtle is following the fish"} and \emph{``The fish is following the turtle"} because their dependency trees differ only in the edge labels of which the model is unaware. Typed DT-LSTM model predicts a higher score of 4.81 for the first sentence and 4.57 for the second. These scores show that the proposed model can semantically differentiate the sentences.

For the second query \emph{``a boy is running towards the ocean"}, the sentence with the most number of word overlap and a matching syntactic structure in the dataset was  \emph{``a couple is running towards the ocean"}. Our model ranked this fifth with a similarity score of $4.0$. The sentence \emph{``a boy is running through the sand"}, which is semantically closer, was ranked first with a score of $4.51$, which indicates that the model considers a change in the subject word as more differentiating. The model can recognize the semantic similarity between the phrases \emph{``towards the ocean"} and \emph{``through the sand"}. 
%The DT-LSTM also gives similar scores for these sentences.
For the third query, all the three retrieved sentences are semantically equivalent, and so are their predicted scores. %The scores of DT--LSTM slightly vary for the three sentences.
\begin{table} 
\centering
\begin{tabular}{@{}p{4.5cm}p{4.5cm}p{1cm}p{1cm}@{}}
\hline \hline
Sentence 1 & Sentence 2 & S & G\\\hline
\hline
The elephant is being ridden by the woman & The woman is riding the elephant & 4.9 & 4.8\\
\hline
Two kids are pulling an inflatable crocodile out of a pool & Two kids are pushing an inflatable crocodile in a pool & 4.23 & 3.7 \\
\hline
A man is riding a horse & A woman is riding a horse & 3.57 & 3.7 \\\hline
A man is preparing some dish & A man is preparing a dish & 4.52 & 5 \\\hline
A man is sitting near a flower bed and is overlooking a tunnel & Two people are sitting on a bench & 1.91 & 1.73\\
\hline
Oil is being poured into the pan by the woman & The boy is playing the piano & 1.09 & 1.0 \\\hline
\end{tabular}
\caption{Sample sentences of different similarity scores from SICK test set data. S is the predicted score and G is the ground truth.}
\label{tab:exsent}
 \end{table}

Table \ref{tab:exsent} lists some sample sentences from the SICK dataset along with the actual score (G) and the predicted score (S). The sentence pairs that were assigned the maximum score by the Typed DT-LSTM model, i.e., 4.9, were active-passive sentence pairs with the same meaning (Example 1 in Table \ref{tab:exsent}). These results show that the model is insusceptible to change in the voice of sentences. We also examined sentence pairs, which differed only by a single word. The changes in subject or object word affected the predicted similarity scores more than the changes in other grammatical roles like determiner or adjectives. In third example, changing \emph{man} to \emph{woman} in the sentence \emph{``A man is riding a horse"} decreased the similarity score to $3.57$, while changing \emph{a} to \emph{some} in the sentence \emph{``A man is preparing a dish"} had little effect on scores. The above examples suggest that the effect of each grammatical relation in the overall meaning composition can be determined only by a detailed investigation of the magnitude of typed dependency embeddings (Refer Section \ref{sec:tdembd}).
%These motivate to conduct further analysis of the learned typed dependency embedding, in section \ref{sec:tdembd}.
\begin{table} 
\centering
\begin{tabular}{@{}p{4.5cm}p{4.5cm}p{1cm}p{1cm}@{}}
\hline \hline
Sentence 1 & Sentence 2 & S & G\\\hline
\hline
A person is doing a trick on a snowboard & A snowboarder is grinding down a long rail & 2.43 & 4.0\\
\hline
A basketball player is on the court floor and the ball is being grabbed by another one & Two basketball players are scrambling for the ball on the court & 2.7 & 4.7 \\
\hline
A person is performing tricks on a motorcycle & The performer is tricking a person on a motorcycle & 3.94 & 2.6 \\\hline
A group of five old adults are lounging indoors & A group of five young adults are lounging indoors & 4.64 & 3.4 \\\hline
A dog is jumping on a diving board & A dog is bouncing on a trampoline & 4.12 & 2.9\\
\hline
\end{tabular}
\caption{List of sentence pairs from SICK test dataset where the predicted score S differed the most from the ground truth G.}
\label{tab:errors}
 \end{table}
 
In Table \ref{tab:errors}, we examine sentence pairs where the predicted score differed the most from human rating. For the first two sentence pairs, the predicted scores are lesser than the ground truth. The model fails to recognize some obscure semantic relations like  \emph{grinding down a long rail} is a \emph{trick} done by a snowboarder, and \emph{ball being grabbed by another one} is the same as  \emph{scrambling in a basketball court}. These mispredictions hint that capturing infrequent senses into a vector encoding is still a challenge.  For the other three pairs, the model predicts higher relatedness scores than human ratings.

\subsection{Typed Dependency Embeddings}{\label{sec:tdembd}}

As explained in section \ref{sec:tydtree}, the relation gate of Typed DT-LSTM learns a task-specific embedding of the Typed dependencies. 
\begin{table}  
\centering
\begin{tabular}{@{}p{3cm}p{3.5cm}p{3cm}p{1cm}@{}}
\hline \hline
Dependencies & Examples& Notation& $M$\\\hline
Direct object & Chef \underline{cooked} the \underline{food} & dobj(cooked,food)& 9.16\\
Nominal modifier  & Food was \underline{cooked} by the \underline{chef} & nmod(cooked,chef)&7.62\\
Adjectival modifier & Sam eats \underline{red} \underline{meat}  & amod(meat,red)& 7.31\\
Nominal subject & \underline{Chef} \underline{cooked} the food & nsubj(cooked,chef)&7.27\\
Conjunct & Bill is \underline{big} and \underline{honest} & conj(big,honest)&6.97\\
Negation modifier & Bill is \underline{not} a \underline{scientist} & neg(scientist,not)&6.90\\
Case marking & I saw a cat \underline{in} the \underline{hat} & case(hat,in)&6.76\\
\hline
\end{tabular}
\caption{List of most significant Universal Dependencies ranked by the model in descending order of their magnitude $M$}
\label{tab:tdranks}  
 \end{table}

We compared the magnitudes of typed dependency embeddings to understand how much each one contributes to the overall meaning composition. Table \ref{tab:tdranks} lists the typed dependencies in the decreasing order of their magnitudes. We find that this order of precedence is in agreement with our human intuition. The relations like direct object($dobj$), nominal modifier($nmod$), adjective modifier($amod$), and nominal subject($nsubj$) make more significant contributions in meaning understanding while relations like goes-with($goeswith$) and adjectival clause($acl$) contribute the least. Surprisingly, the pseudo dependency ``root" didn't top the list.

\begin{figure}[!h]
  \centering
    \includegraphics[width=0.9\linewidth]{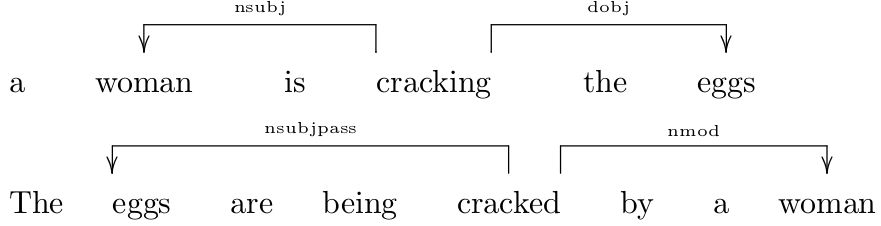}
    \caption{Dependency trees of two semantically equivalent sentences; one in active and other in passive voice}
  \label{fig:crack}
\end{figure}

A detailed analysis of these embedding reveals some interesting observations.
Consider two sample sentences expressing the same meaning, one in active voice and the other in passive:\\
(1) A woman is cracking the eggs.\\
(2) The eggs are being cracked by a woman.\\
Some typed dependencies of the two sentences are shown in Figure \ref{fig:crack}.
\begin{comment}
\xytext{
  \xybarnode{a} &~~~~&
  \xybarnode{woman}
    &~~~~~~&
  \xybarnode{is} &~~~~&
  \xybarnode{cracking}
      \xybarconnect[3](UR,U){4}"^{\tiny dobj}"
      \xybarconnect[3](UL,U){-4}"_{\tiny nsubj}"
     &~~~~~~&
  \xybarnode{the} &~~~~&
  \xybarnode{eggs}
}
\\
\xytext{
  \xybarnode{The} &~~&
  \xybarnode{eggs}
    &~~~&
  \xybarnode{are} &~~&
  \xybarnode{being}
     &~~~&
  \xybarnode{cracked}
      \xybarconnect[3](UR,U){6}"^{\tiny nmod}"
      \xybarconnect[3](U,UL){-6}"_{\tiny nsubjpass}"
     &~~~&
  \xybarnode{by}
    &~~~&
    \xybarnode{a} &~~&
  \xybarnode{woman}
     &~~~&
}
\end{comment}

In the first sentence, typed dependency between the word pair (cracking, woman) is $nsubj$ and that between (cracking, eggs) is $dobj$. In the second sentence, the dependency types for the two pairs are $nmod$ and $nsubjpass$, respectively. We found that the vector difference (both in magnitude and direction)  between $nsubj$ and $nmod$ is approximately the same as that between $dobj$ and $nsubjpass$ which implies,
\begin{center}
$nsubj:nmod :: dobj:nsubjpass$ 
\end{center}
This similarity suggests that our model has learned not just the relationship types in word pairs but also how these relationship types change as they transfer from active to passive voice. 
\begin{table}
\centering
%\caption{Binary Sentiment classification.}
\begin{tabular}{@{}p{4cm}p{2cm}@{}}
\hline \hline
Model & Accuracy(\%)
\\\hline
LSTM  & 84.9\\
Bi- LSTM & 87.5  \\
2-layer LSTM & 86.3\\
2-layer Bi-LSTM & 87.2 \\
\hline
CT-LSTM & 88.0\\
DT-LSTM & 85.7\\
\hline
\textbf{Typed DT-LSTM} & \textbf{86.9}\\\hline
\end{tabular}
\caption{Comparison of Typed Dependency Tree-LSTM with LSTM models for binary sentiment classification on SST dataset. The values are taken from previously published results.}
\label{tab:results1}
\end{table}
\subsection{Modeling sentiment analysis}
We compared the accuracy of the Typed DT-LSTM model with that of other state-of-the-art LSTM models (Table \ref{tab:results1}) on the task of predicting the sentiment of phrases in movie reviews. We found that the proposed model outperformed standard LSTM as well as DT-LSTM in the accuracy of prediction. The accuracy is comparable with that of bidirectional LSTM but lesser than that of the constituency tree-LSTM (CT-LSTM). %Given the fact that data available for the training dependency tree models were much lesser compared to the constituency tree model, it is not reasonable to conclude constituency trees are more appropriate structures for capturing sentiments.

\begin{figure*}[t!]
    \centering
    \begin{subfigure}[t]{0.45\textwidth}
        \centering
        \includegraphics[scale=0.35]{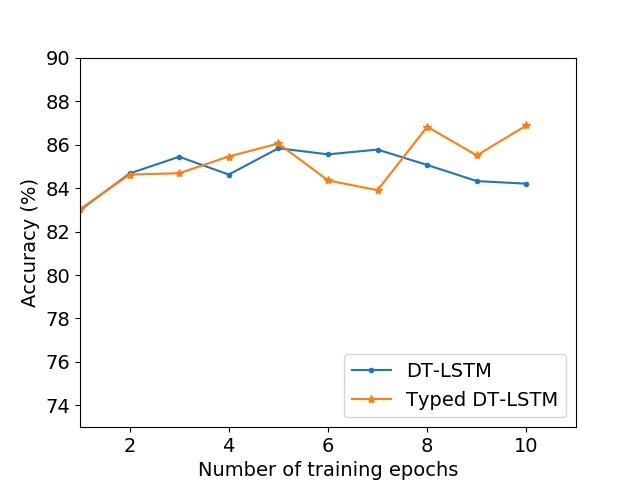}
        \caption{Accuracy}
    \end{subfigure}%
    \begin{subfigure}[t]{0.5\textwidth}
        \centering
        \includegraphics[scale=0.35]{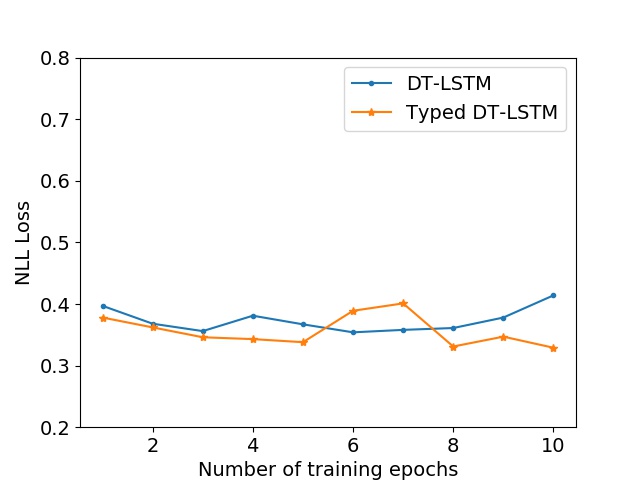}
        \caption{Negative Log Likelihood (NLL) Loss}
    \end{subfigure}
    \caption{Comparison of learning curves of DT-LSTM and Typed DT-LSTM model for Sentiment Classification Task using (a) Accuracy (\%) and (b) Negative Log Likelihood (NLL) loss. Learning curves are calculated from Stanford Sentiment Treebank (SST) test dataset.}
    \label{fig:senticurves}
\end{figure*}

Typed DT-LSTM model obtains a Precision of 88.4\%, Recall of 84.8\%, and F1--score of 86.6\% in identifying positive sentiment on the test dataset. The Precision, Recall, and F1-score of DT-LSTM are 87.62\%, 82.28\%, and 85.39\%, respectively. The results clearly indicate that Typed LSTM is superior to DT-LSTM in recognizing sentiments of text. Figure \ref{fig:senticurves} plots the learning curves of Typed DT-LSTM and the base model DT-LSTM for sentiment classification. Typed DT-LSTM has nearly the same learning rate as DT-LSTM but outperforms it after the seventh epoch. The optimal hyperparameters\ref{tab:hyperp} for the model are chosen empirically. 
Table \ref{tab:sentiments} shows some randomly selected examples from the SST test dataset along with the predicted and actual sentiment labels categorized into four; true negatives, true positives, false positives, and false negatives. The first and second categories list reviews which the model labels correctly. Though these sentences had positive words like \emph{thank}, \emph{glad}, and \emph{inspire}, the model accurately detects the overall polarity as negative. Similarly, in the positive review, the negative words do not mislead the model into assigning wrong labels. But in the third category, the sentences containing phrases of opposite polarity are misclassified as positives. The reviews in the fourth category are rated negative though they express a positive sentiment. Sentiment classification of longer text with fairly complex structures is still a challenge to be addressed.

\begin{table} 
\centering
\begin{tabular}{@{}p{11cm}p{0.65cm}p{0.65cm}@{}}
\hline \hline
Input sentences & S & G\\\hline
\hline
%Usually when I get this much syrup, I like pancakes to go with it.& 0 & 0 \\
I wish I could say ``Thank God it's Friday'', but the truth of the matter is I was glad when it was over.& 0 & 0 \\
Maybe it's asking too much, but if a movie is truly going to inspire me, I want a little more than this. & 0 & 0 \\
\hline
Both deeply weird and charmingly dear.& 1 & 1 \\
%The film is just a big, gorgeous , mind-blowing , breath-taking mess.& 1 & 1 \\
%Sharp, lively, funny and ultimately sobering film.& 1 & 1 \\
A provocative movie about loss, anger, greed, jealousy, sickness and love. & 1 & 1 \\
\hline
Begins as a promising meditation on one of America's most durable obsessions but winds up as a slender cinematic stunt.& 1 & 0 \\
%Proof that a thriller can be sleekly shot, expertly cast, paced with crisp professionalism... and still be a letdown if its twists and turns hold no more surprise than yesterday's weather report.& 1 & 0 \\
A long slog for anyone but the most committed Pokemon fan. & 1 & 0 \\
\hline
Whether you're moved and love it, or bored or frustrated by the film, you'll still feel something.& 0 & 1 \\
If no one singles out any of these performances as award-worthy, it's only because we would expect nothing less from this bunch.& 0 & 1 \\
%The best thing the film does is to show us not only what that mind looks like, but how the creative process itself operates.& 0 & 1 \\
\hline
\hline
\end{tabular}
\caption{Some examples from the Stanford Sentiment Treebank (SST) test dataset along with the predicted score S and the gold label G. 0 represents negative, and 1 represents positive sentiment polarity.}
\label{tab:sentiments}
 \end{table}
 
\subsection{Theoretical and Practical Implications of Research}
From a theoretical point of view, our research findings imply the significance of grammatical relations in modeling sentence semantics. Most existing deep learning researches in NLP have focused only on the word meaning and syntactic structure of sentences. Hence, those sentence representations are insufficient to recognize semantic differences due to changes in the grammatical role of their words. The proposed models would serve as a base for future studies on role of typed dependencies in meaning understanding.

This research has practical implications in NLP tasks like Question Answering, Natural Language Inference, Recognizing Textual Entailment, Paraphrase Identification, and Duplicate Question merge, as semantic modeling is a fundamental and crucial step in all of these. Tree-LSTMs has also recently gained much attention in biomedical literature mining for extracting relations like protein-protein interaction, chemical-gene association. So the proposed enhancement in the Tree-LSTM model would advance recent research efforts in these directions as well.
\section{Conclusion and Future work}{\label{sec:con}}
The contributions of this research are three-fold. First, a Relation gated LSTM (R-LSTM) generic architecture has been proposed. The relation gate of R-LSTM network learns separate gating vectors for each type of relationship in the input sequence. As a second contribution, a Typed Dependency Tree-LSTM model has been proposed that make use of dependency parse structure and grammatical relations between words for sentence semantic modeling. Third, a qualitative analysis of the role of typed dependencies in language understanding is performed. Experiments show that the proposed model outperforms DT-LSTM in terms of performance and learning speed in semantic relatedness scoring tasks and sentiment analysis.

We compared the proposed model with other state-of-the-art methods for semantic composition. The Typed DT-LSTM can identify subtle relation between phrases in the sentence pair, and thereby score better correlation with the human rating. 

%To the best of our knowledge, this is the first deep learning model that has efficiently employed universal dependencies in Tree-LSTMs for sentence modeling.
The role of dependency types in semantic composition has not yet been explored in the deep learning context. The proposed computational models would encourage future research efforts in this direction. We intend to experiment with Typed DT-LSTM for modeling sentence semantics in other NLP tasks like paraphrase detection, natural language inference, question answering, and image captioning where DT-LSTMs has already shown promising results.

A detailed analysis of the typed dependency embeddings learned by the model reveals some interesting insights into language understanding. From a linguistic point of view, these embeddings are worth exploring further.

Finally, the relation gated LSTM architecture proposed in this work is an idea that can be adapted to other domains as well. LSTMs has become de-facto for many sequence modeling problems. For those tasks that need to model not just the nodes but also the different kinds of links between them, R-LSTM would be an alternative to LSTMs. Our results show that R-LSTMs are capable of learning the relation between LSTM units. The benefits of using relation gate in other gated architectures like Tree-GRU are also worth exploring.

%\section*{References}

\bibliography{tydtreelstm}

\end{document}